\documentclass[10pt,twocolumn,letterpaper]{article}

\usepackage{wacv}
\usepackage{times}
\usepackage{epsfig}
\usepackage{graphicx}
\usepackage{amsmath}
\usepackage{amssymb}
\usepackage{color}
\usepackage{comment}
\usepackage{algorithm}
\usepackage[noend]{algpseudocode}
\usepackage{verbatim}
\usepackage{cite}

\usepackage[pagebackref=true,breaklinks=true,letterpaper=true,colorlinks,bookmarks=false]{hyperref}

\wacvfinalcopy 

\def\wacvPaperID{35} 

\ifwacvfinal\fi
\begin{document}

\newcommand{\note}[1]{\textcolor{red}{NOTE: #1}}
\newcommand{\idea}[1]{\textcolor{blue}{IDEA: #1}}
\renewcommand{\baselinestretch}{0.96} 

\title{\vspace{-0.25in}
Style Transfer for Light Field Photography
\vspace{-0.05in}
}

\author{David Hart, Jessica Greenland, and Bryan Morse \\
Brigham Young University\\
{\tt\small \{davidmhart, jessica.greenland, morse\}@byu.edu}
}

\maketitle
\ifwacvfinal\thispagestyle{empty}\fi

\begin{abstract}
As light field images continue to increase in use and application, it becomes necessary to adapt existing image processing methods to this unique form of photography. 
In this paper we explore methods for applying neural style transfer to light field images. 
Feed-forward style transfer networks provide fast, high-quality results for monocular images, but no such networks exist for full light field images. 
Because of the size of these images, current light field data sets are small and are insufficient for training purely feed-forward style-transfer networks from scratch. 
Thus, it is necessary to adapt existing monocular style transfer networks in a way that allows for the stylization of each view of the light field while maintaining visual consistencies between views.
To do this, we first generate disparity maps for each view given a single depth image for the light field.
Then in a fashion similar to neural stylization of stereo images, we use disparity maps to enforce a consistency loss between views and to warp feature maps during the feed forward stylization.
Unlike previous work, however, light fields have too many views to train a purely feed-forward network that can stylize the entire light field with angular consistency. 
Instead, the proposed method uses an iterative optimization for each view of a single light field image that backpropagates the consistency loss through the network. 
Thus, the network architecture allows for the incorporation of pre-trained fast monocular stylization network while avoiding the need for a large light field training set.
\end{abstract}
\vspace{-0.125in}

\section{Introduction}
\label{sect:introduction}

Light field photography continues to be a technology that presents many challenges and problems to overcome including memory constraints and editing difficulties. 
However, it also presents fascinating capabilities that are not possible with regular images, such as novel view synthesis and focal stack generation. 
In recent years, we have seen light field technology adopted for more and more commercial applications, from virtual- and augmented-reality systems to dedicated light-field cameras such as those from Lytro (based on the work of~\cite{NgDissertation}). Many variants on multiple-camera imaging configurations are beginning to become more commonplace in the commercial market (ex. dual-camera configurations increasingly found in cell phones), and methods for working with these images often draw from concepts in light-field literature.

\begin{figure}[t]
\begin{center}
\includegraphics[width=0.80\linewidth]{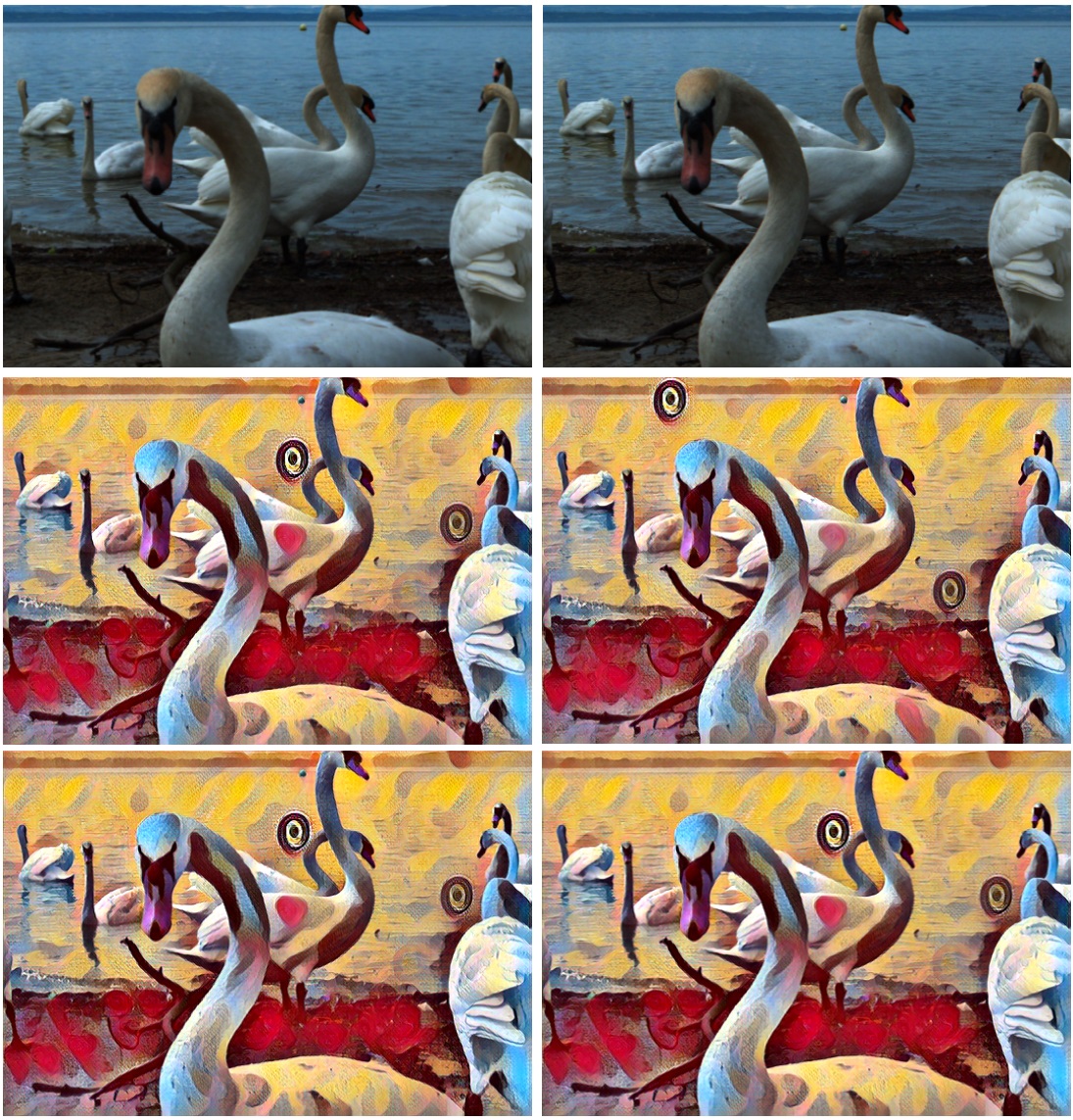}
\vspace{-0.1in}
\end{center}
   \caption{Stylization of two views from a light field with 81 images. 
   Even though there are minimal visual differences between these two views from the same light field image (top), the stylization of these two views (middle) results in dramatic differences in the coloring and features that are not present in the original.
   Our proposed method results in consistency not only between these two views (bottom) but the entire set of 81 views.
   \vspace{-0.125in}
   }
\label{fig:naive}
\end{figure}

``Painterly'' and other forms of non-photorealistic rendering of one image based on the style of another have a long history in computer graphics.
We refer the interested reader to the original work of~\cite{Hertzmann:2000:PRV:340916.340917,PainterlyRendering,ImageAnalogies} as well as more recent work in~\cite{Lu:2010:IPS} and surveys in~\cite{PainterlySurvey2012,Hertzmann:2018aa}.
This area has seen a resurgence in recent years due to the application of deep neural networks to the problem.
Rather than trying to analyze brush strokes, texture, or other properties explicitly, recent methods for {\em neural style transfer} treat the problem as one of optimizing for preservation of the content of one image and the stylistic properties of another.
The groundbreaking work of Gatys \etal~\cite{OriginalStyleTransfer} and the feed-forward method presented by Johnson \etal~\cite{FeedForwardStyleTransfer} have opened the door to many variations on these ideas, including improved methods of direct optimization on the resulting image~\cite{Kolkin2019,Puy2019}, modifications to the feed-forward stylization network~\cite{Kotovenko2019,Li2017,Li2018}, stylizing for texture synthesis~\cite{Ulyanov2016,Ulyanov2017}, using depth information to inform stylization~\cite{DepthStyleTransfer}, training a single network to perform multiple stylizations~\cite{ConditionalStyleTransfer}, providing greater control over the stylization~\cite{Gatys2017a}, stylizing video~\cite{Chen_2017_ICCV,Gupta_2017_ICCV,Huang_2017_CVPR,Ruder:2018aa}, and stylizing stereo pairs~\cite{StereoStyleTransfer1,StereoStyleTransfer2}.

We seek to expand neural stylization to light-field images. 
This would present new possibilities in stylization not seen before, such as novel viewpoint generation and dynamic refocusing of stylized images.
Expanding stylization to different types of photography has been done before for 360$^\circ$ video~\cite{Ruder:2018aa}, RGB-D~\cite{DepthStyleTransfer}, and stereo imaging~\cite{StereoStyleTransfer1, StereoStyleTransfer2}, but light fields present their own challenges that cannot be solved by simply generalizing one of these methods.

The naive approach to stylizing a light field would be to use a single-image style transfer network for each view independently, but such a network has no notion of angular consistency between the images, which generally leads to visual differences as shown in Fig.~\ref{fig:naive}.  
Both~\cite{StereoStyleTransfer1} and~\cite{StereoStyleTransfer2} note these inconsistencies when naively stylizing stereo images (what~\cite{StereoStyleTransfer1} refers to as the ``baseline'' method). This problem is only exacerbated when one goes from two images for a stereo pair to the much larger number of views in a light-field image (typically on the order of 50--200).
This inconsistency degrades the scene geometry that could normally be calculated from a light field. 
Without this intrinsic consistency and geometry, none of the effects that are normally associated with light fields can be processed in a visually coherent manner.

This paper presents a method for stylizing light field images in a way that maintains angular consistency between the different views.
We first demonstrate an effective way to generate disparity maps for each view of the light field given only a single depth map.
Then, we exploit the scene geometry to inform the network of locations that are the same between views, extending the concept of multiple-image consistency used previously for video sequences~\cite{Ruder:2018aa} and stereo pairs~\cite{StereoStyleTransfer1,StereoStyleTransfer2}.
Additionally, the proposed method does not require retraining the base feed-forward style transfer network~\cite{FeedForwardStyleTransfer} specifically for light fields. 
This allows previously trained networks to be used and avoids the need for large light-field datasets.

As light field technology continues to improve and be adopted for more applications, the need for methods of light field editing will continue to grow, especially as the capabilities of image-processing neural networks also continue to expand. 
Although this work is specific to neural style transfer, it potentially lays a foundation for light-field consistency optimizations that could generalize to other applications.


\section{Related Work}
\label{sect:related}


Light field research continues to expand as light field cameras become increasingly available for commercial applications. 
Light fields can be used to create novel views and generate focal stacks~\cite{Lumigraph,FirstLightField,NgDissertation}. 
Light fields can also be used to calculate more accurate depth estimates using epipolar images and light field features~\cite{LightFieldDepth1, LightFieldDepth2, LightFieldDepth3, LightFieldDepth4}. 
The multiple angles and views of a light field also allow for separation of the diffuse and specular components of reflectance~\cite{LightFieldIntrinsics1, LightFieldIntrinsics2}. 
Work has even been done to use light fields for classification, especially of materials~\cite{wang2016dataset}.

As described in the introduction, the work in this paper seeks to extend the ideas of neural stylization to light field images, for which the key challenge is maintaining angular consistency between the multiple stylized views.
Maintaining such consistency in the result is an essential element of any approach that edits multiple images with corresponding content, such as video sequences or stereo pairs~\cite{Luo:2015aa,Morse:2012fk}.
The key in these approaches is to identify or use existing methods to identify correspondences between the images (optical flow for video, stereo correspondence, etc.) and ensure that the results maintain this correspondence.

Ruder \etal~\cite{Ruder:2018aa} first introduced the idea of using such correspondences to extend neural style transfer to video sequences.
They used optical flow to identify the correspondences and extended the optimization-based stylization approach of Gatys \etal~\cite{OriginalStyleTransfer} to include an additional consistency loss term.
These ideas were extended by Chen \etal~\cite{Chen_2017_ICCV}  to train a feed-forward network (building on~\cite{FeedForwardStyleTransfer}) to produce similarly consistent video stylization.

Chen \etal~\cite{StereoStyleTransfer2} and
Gong \etal~\cite{StereoStyleTransfer1}  
have each proposed methods for photo-consistent style transfer for stereo pairs, which can be thought of as a much smaller subset (two images) of a light field. 
The approach of Chen \etal~\cite{StereoStyleTransfer2} builds on a network structure similar to their earlier video-stylization work~\cite{Chen_2017_ICCV} to train a feed-forward network to learn to perform the stylization.
Gong \etal~\cite{StereoStyleTransfer1} likewise train a feed-forward network to perform stylization.

This paper incorporates elements of both~\cite{StereoStyleTransfer2} and~\cite{StereoStyleTransfer1}, but neither of these methods for stylizing image pairs directly generalize to the much larger number of views in light fields because 1)~both methods rely on having pairwise disparity maps from each view to the other, 2)~both depend on a single network to stylize all views, and 3)~both rely on retraining the network on a large dataset.
Extrapolating such an approach to a full light field is simply not viable.



\section{Multiview Angular Consistency}
\label{sect:masks}

To enforce angular consistency between multiple views, pixel-wise correspondence needs to be determined for each view in the light field.
The most effective way of doing this is by using the depth map that is calculated from the epipolar images of the light field (e.g.,~\cite{LightFieldDepth1,LightFieldDepth2,LightFieldDepth3,LightFieldDepth4}), leveraging more information from the field than in two-image stereo correspondence.
%
Using such methods, the depth map is generally precomputed for light field images in standard datasets~\cite{LFDatasets1,LFDatasets2} and is easily accessible.
However, such methods usually produce a depth map only for the central (reference) view and not for each separate view in the light field~\cite{Johannsen2017}, which must be addressed for consistent stylization.
The method proposed here is independent of the choice of method used to estimate depth and assumes that the depth map for the central view has been precomputed.

This paper adopts the notation of~\cite{LightFieldDepth1} and most other recent work by indexing the subaperture views by $(s,t)$ and the pixels within each view by $(x,y)$.
Individual subaperture views are thus denoted as $I_{s,t}$ with the central image as $I_{0,0}$ and others indexed using both positive and negative relative $(s,t)$ indices.

Although the central-view depth map is often not calibrated, it provides relative scene geometry and can be inverted and calibrated to produce a pixel disparity map $D_{0,0}$ using a simple optimization algorithm to estimate the unknown scaling due to focal length, imaging pixel density, and the (effective) baseline separation of the subaperture views~\cite{diebold-etal-2015-SPIE}.
Specifically, this optimization inversely scales the depth map to produce the disparity map $D_{0,0}$ that maximizes the correspondence between the central view and the adjacent view to the right, giving us the mapping $I_{0,0} \rightarrow I_{1,0}$.
For many light fields, including those shown in our results, there is also an additional translation and cropping for each view, resulting in a planar horopter at an unknown depth and a mix of both positive and negative disparities.
To accommodate such cases, we add a second optimized calibration parameter that adds a translation bias.
This allows for negative disparities even though the inverted depth map is all positive values.

Because stereo images are typically separated along a horizontal baseline, disparity is often mistakenly thought of solely as the degree of opposite horizontal movement as one moves in a horizontal direction.
But it is important to remember that disparity is the degree of apparent opposite movement as one moves the camera in {\em any} direction.
Thus, the reference disparity map $D_{0,0}$ thought of as horizontally mapping  $I_{0,0} \rightarrow I_{1,0}$ can just as easily be used to provide the mapping $I_{0,0} \rightarrow I_{0,1}$ as one moves vertically.
Similarly, the vector field that maps $I_{0,0} \rightarrow I_{s,t}$ can be calculated using $D_{0,0}(x,y) \ [s,t]^T$.

As noted previously, depth maps for light fields are often computed only for the central (reference) view, allowing computation of a disparity map for this view only.
A disparity map for an image allows for forward-mapping of each pixel to where it maps to in another view, which can be many-to-one in the case of occlusion or none-to-one in the case of disocclusion.
Instead of using forward warping, however, we desire to use backward warping of the central view to the other views, which requires disparity maps $D_{s,t}(x,y)$ for each of the other views.

\begin{figure}[t]
\begin{center}
\includegraphics[width=0.75\linewidth]{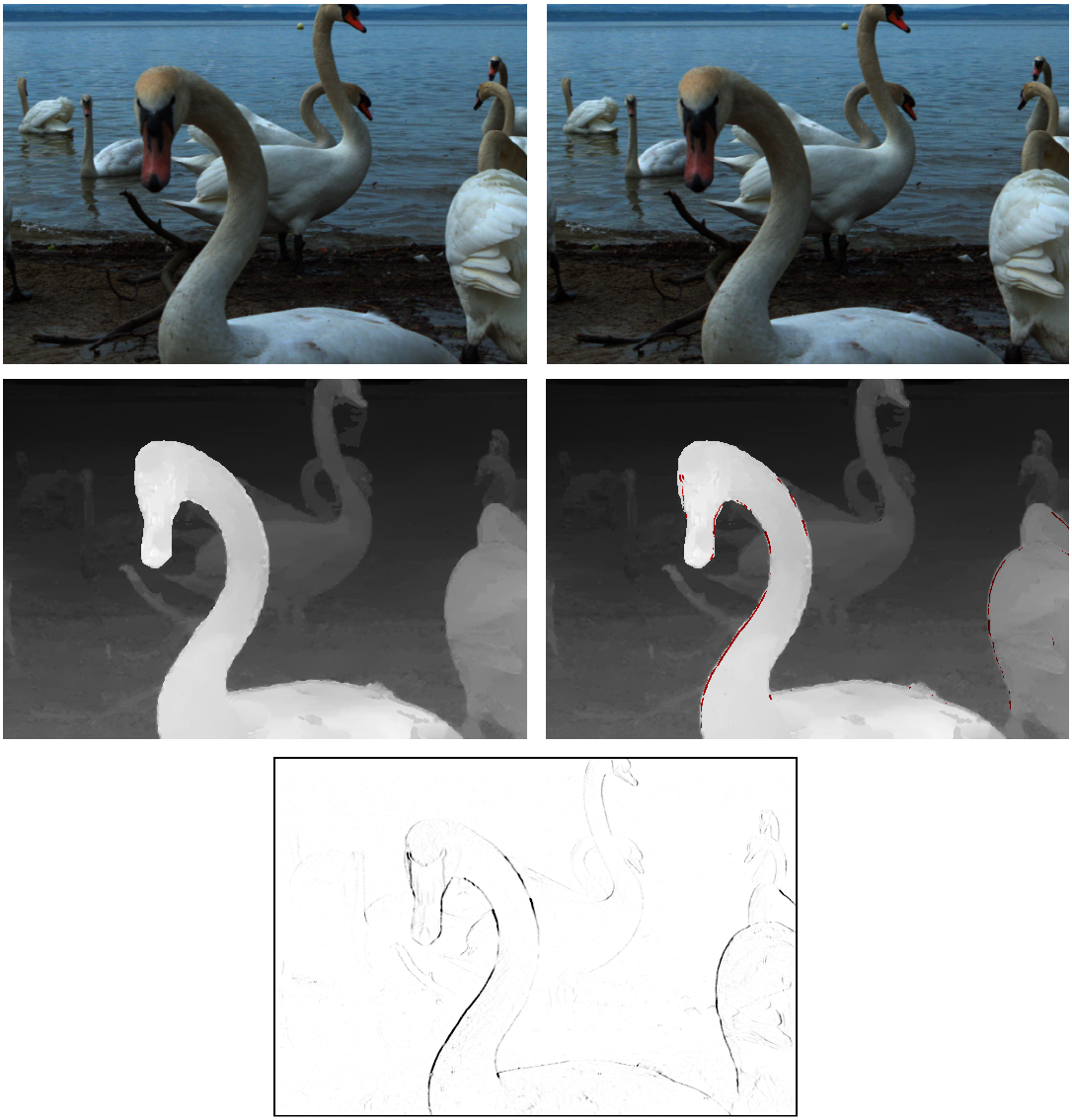}
\vspace{-0.1in}
\end{center}
   \caption{
    Reversing the central disparity map $D_{0,0}$ to produce (partial and masked) disparity maps $D_{s,t}$ for other viewpoints.
    Top: Views $I_{0,0}$ and $I_{2,0}$ of the light field. 
    Middle: The disparity map from $I_{0,0} \rightarrow I_{2,0}$ (left) and the reversed disparity map from $I_{2,0} \rightarrow I_{0,0}$ (right). 
    Red denotes areas of occlusion that are not seen in the central view. 
    Bottom: The consistency mask $M_{s,t}$ with fuzzy values for partial occlusions or low-confidence correspondences, with zero (black) for points with no correspondence.
   \vspace*{-0.15in}
   }
\label{fig:Disparity}
\end{figure}

We assign disparity $D_{s,t}(x,y)$ for each pixel in each view through a simple search to find the set of pixels (potentially empty, one, or more than one) in the central image that map {\em to} that pixel $(x,y)$ in view $(s,t)$.
While this might seem to be an expensive search, it can be constrained in multiple ways: 
1)~epipolar geometry constrains the corresponding points to match along the line 
$(x - s \ D_{s,t}(x,y), y - t \ D_{s,t}(x,y))$, 
and
2)~the minimum and maximum disparities in the central disparity map $D_{0,0}$ can be used to bound the search range along, or near, the epipolar line.
%
%
For each candidate matching point $(x',y')$, we consider all candidate matches that satisfy
\begin{equation}
\small
    \left\| 
        (x' + s \ D_{0,0}(x',y'), y' + t \ D_{0,0}(x',y')) - (x,y)
    \right\| 
    < \epsilon 
    \label{eq:disparity-symmetry}
\end{equation}
for some small value of $\epsilon$ large enough to account for discrete pixel sampling.  (We use $\epsilon = 1.4$.)

\begin{figure*}[t]
\begin{center}
\includegraphics[width=0.65\linewidth]{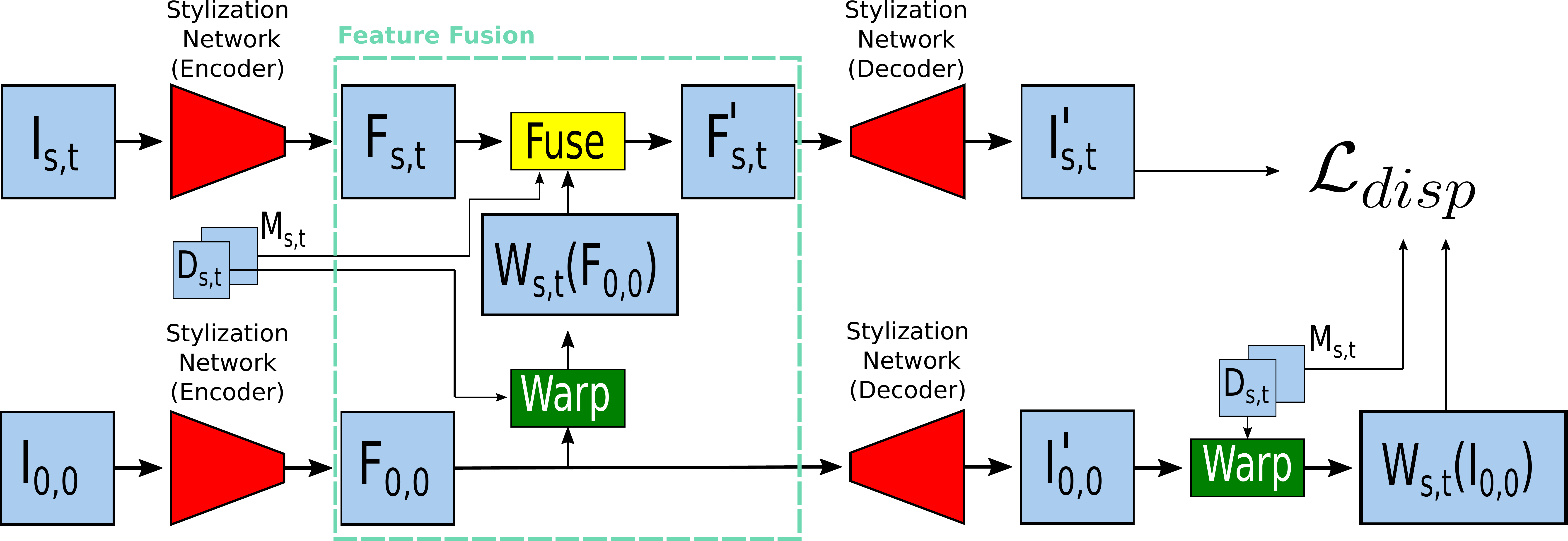}
\vspace{-0.1in}
\end{center}
   \caption{
   Network architecture for neural stylization of light field images. The lower channel stylizes $I_{0,0}$ to produce $I'_{0,0}$ with intermediate feature map $F_{0,0}$, all of which are then held fixed.
   The upper channel is repeated for each other view $I_{s,t}$.
   The encoded feature map for this image $F_{s,t}$ is then fused with the warped $F_{0,0}$ using the confidence map $M_{s,t}$ and decoded to produce $I'_{s,t}$.
   The masked disparity loss between $I'_{s,t}$ and a warped $I'_{0,0}$ is calculated and is backpropagated through the entire network updating only the encoding and decoding parts of the pre-trained feed-forward stylization subnetwork.
   This process is repeated for some number of epochs for each view $I_{s,t}$ to optimize the result.
   \vspace{-0.05in}
   }
\label{fig:BigPicture}
\end{figure*}

Using the idea of stereo symmetries and plausible disparities from~\cite{Sun:2005uq}, we then select the potentially matching candidate with the largest disparity $D_{0,0}(x',y')$, which ensures that the front-most surface is chosen when occlusion causes a many-to-one forward mapping for the point.
If no satisfactory match is found, this indicates the disocclusion that would result in a none-to-one forward mapping.
An example of inverting the central disparity map can be found in Fig.~\ref{fig:Disparity}, with ``no correspondence'' disoccluded regions indicated in red.

During this search we simultaneously compute a correspondence confidence map $M_{s,t}$ (as also shown in Fig.~\ref{fig:Disparity}) where $M_{s,t}(x,y) \in [0,1]$ is determined by comparing the quality of the pixel correspondences (using normalized RGB distance) determined through the just-described search process:
\begin{equation}
    \small
    M_{s,t}(x,y) = 1 - ||I_{s,t}(x,y) - W(I_{0,0},D_{s,t})(x,y)||/\sqrt{3}
    \label{eq:mask}
\end{equation}
where $W(I_{0,0},D_{s,t})$ denotes the backward warping from image $I_{0,0}$ based on the disparity map $D_{s,t}$:
\begin{equation}
    \footnotesize
    W(I_{0,0},D_{s,t})(x,y) = \hat{I}_{0,0}(x - s \ D_{s,t}(x,y), y - t \ D_{s,t}(x,y))
\end{equation}
with $\hat{I}$ denoting interpolation of image $I$
and all pixel values assumed to be in the range $[0,1]$ for each color channel.
If no correspondence was found through the search using Eq.~\ref{eq:disparity-symmetry}, the backward warping is undefined and $M_{s,t}(x,y)$ is set to 0.
We use this confidence map as a mask when enforcing consistency (similar to the function of the ``gate map'' in~\cite{StereoStyleTransfer1}).
This allows greater inconsistency where the original correspondences are uncertain, when partial-pixel effects near object edges produce imperfect correspondence, or when there is otherwise angular inconsistency (e.g., specular reflections~\cite{wang2017svbrdf}) and not enforcing consistency at all where no correspondence exists.

\section{Style Transfer for Light Fields}
\label{sect:full-method}

Training a feed-forward style-transfer network is a time-consuming process that generally requires thousands or even millions of training images~\cite{FeedForwardStyleTransfer}. 
Since light field images require relatively large amounts of storage compared to typical single or even stereo images, existing light field datasets are very small and not sufficient for the task of training a feed-forward style-transfer network. 
Thus, any consistency constraints that are instituted must work within the framework of existing pre-trained style-transfer networks.

We propose a method for light field style transfer that maintains angular consistency between views.
For this method, we use a pre-trained feed-forward network as described in Johnson \etal~\cite{FeedForwardStyleTransfer}, specifically the implementation found at~\cite{PytorchStyleTransfer}.
While we choose to work with this specific implementation, our method could also be adapted to fit within the structures of more recent feed-forward stylization networks, such as those found in~\cite{Kotovenko2019} and~\cite{Li2018}.
An overview of our architecture is shown in Fig.~\ref{fig:BigPicture}.

To stylize the light field, we first encode the features $F_{0,0}$ of the central image $I_{0,0}$ using the first (encoder) half of the stylization network.
These features are then decoded using the second (decoder) half of the network to produce $I'_{0,0}$.
These are then held fixed as we stylize the rest of the views.

For each other view $I_{s,t}$ of the light field, the features $F_{s,t}$ are also encoded and blended using the correspondence confidence map $M_{s,t}$ with a version of the central-view features warped to view $(s,t)$ using $D_{s,t}$:
\begin{equation}
    F'_{s,t} = 
    M_{s,t} \odot W(F_{0,0},D_{s,t}) + (1 - M_{s,t}) \odot F_{s,t}
    \label{eq:fuse}
\end{equation}
This is similar to the process described in~\cite{StereoStyleTransfer1} (one of the dual channels) and~\cite{StereoStyleTransfer2} (single-directional variant).
The warped-and-fused feature map is then decoded into the stylized image $I'_{s,t}$.

A disparity (angular consistency) loss~\cite{StereoStyleTransfer2,StereoStyleTransfer1} is calculated using $I'_{s,t}$ and a warped version of $I'_{0,0}$, again modulated by correspondence confidence map $M_{s,t}$:
\begin{equation}
    \mathcal{L}_{disparity} = ||M_{s,t}\odot(I'_{s,t}-W(I'_{0,0},D_{s,t} ))||^2 .
    \label{eq:disparity-loss}
\end{equation}
\noindent The disparity loss is then backpropagated through the network. 
This repeats until convergence or a maximum number of iterations is reached.

This process is repeated for each subaperture view as described in Algorithm~\ref{alg:style}.
In our implementation, we use a learning rate of 1e-2 and run it for a maximum of 50 epochs (though we found most views converge after 40 epochs approximately).
We also use the overfit stylization network from one view as the initial stylization network for the next view rather than resetting the network.  
We have found that this works best when the shift from one view to the next is small, so we visit the different views $I_{s,t}$ by alternating the 
ordering on successive rows (i.e., boustrophedonically) so that the shift in viewpoint is always to an adjacent view.
We also double the epochs for the first view visited to increase stability of stylization.

\begin{algorithm}
\footnotesize
\caption{Light Field Style Transfer}\label{alg:style}
\textbf{Require}: Pre-trained Style Transfer Network $\theta$ on Image S

\textbf{Require}: Light Field $I$, Disparity Maps $D$, and Consistency Masks $M$

\textbf{Returns}: Style Transferred Light Field $I'$
\begin{algorithmic}[1]
\State $F_{0,0} \gets \theta_{encode}(I_{0,0})$ 
\State $I'_{0,0} \gets \theta_{decode}(F_{0,0})$
\For{$s,t$ in $I$}
\For{$epochs$}
\State $F_{s,t} \gets \theta_{encode}(I_{s,t})$
\State $F'_{s,t} \gets M_{s,t} \odot W(F_{0,0},D_{s,t}) + (1 - M_{s,t}) \odot F_{s,t}$
\State $I'_{s,t} \gets \theta_{decode}(F'_{s,t})$
\State $\mathcal{L}_{disparity} \gets ||M_{s,t}\odot(I'_{s,t}-W(I_{0,0},D_{s,t}) )||^2$ 
\State $\theta \gets $BackProp$(\theta,\mathcal{L}_{disparity})$
\EndFor\label{inner}
\State $F_{s,t} \gets \theta_{encode}(I_{s,t})$
\State $F'_{s,t} \gets M_{s,t} \odot W(F_{0,0},D_{s,t}) + (1 - M_{s,t}) \odot F_{s,t}$
\State $I'_{s,t} \gets \theta_{decode}(F'_{s,t})$
\EndFor\label{outer}
\State \textbf{return} $I'$
\end{algorithmic}
\end{algorithm}


Although our integration of warped and fused feature is similar to the approaches in~\cite{StereoStyleTransfer2} and~\cite{StereoStyleTransfer1}, there are distinct modifications necessary to allow such an approach to work for light fields beyond simply the number of views.

In \cite{StereoStyleTransfer2}, these features are warped to a common hypothetical view that is located halfway between the two images in the stereo pair, and then the two are fused.
In a light field for which only a central-view depth map has been computed, however, the only reference point that can act as a common view for all $N \times N$ images is the central view.
Thus, only the central view features are warped and fused with other view features.
This is done using bilinearly resized versions of the disparity maps and consistency masks to match the resolution of the features.

In \cite{StereoStyleTransfer1}, the features are warped and fused in both directions, providing a more consistent version of the features between the pair.
While this process provides good results for an image pair, it does not generalize to light fields.
Allowing all $N \times N$ views of a light field to have an influence on the central view leads to an aggressively averaged set of features, which leads to very blurry output images.
Thus, in the method proposed here, the central view features are held fixed during all stages of the algorithm.
This restriction on the central view is necessary in order to force the network to converge in a way that keeps consistency between all views of the light field while maintaining high-quality output.

It is important to note that our architecture is not used to train a feed-forward network for light fields. 
Instead, our algorithm provides a method for optimizing a single light field image in a reasonable amount of time. 
It is similar to the optimization presented by Gatys \etal~\cite{OriginalStyleTransfer}, but warping of the features and initializing with feed-forward stylization allows the optimization to converge much faster and does not require training on perceptual loss.


\section{Results and Evaluation}
\label{sect:results}

Since there are no other methods for neural stylization of light fields to compare against, we present qualitative results (visual examples) and quantitative evaluation of the degree to which the resulting stylization preserves both perceptual factors (content and style loss) and inter-view angular consistency (disparity loss).

\begin{figure}[t]
\begin{center}
\includegraphics[width=.75\linewidth]{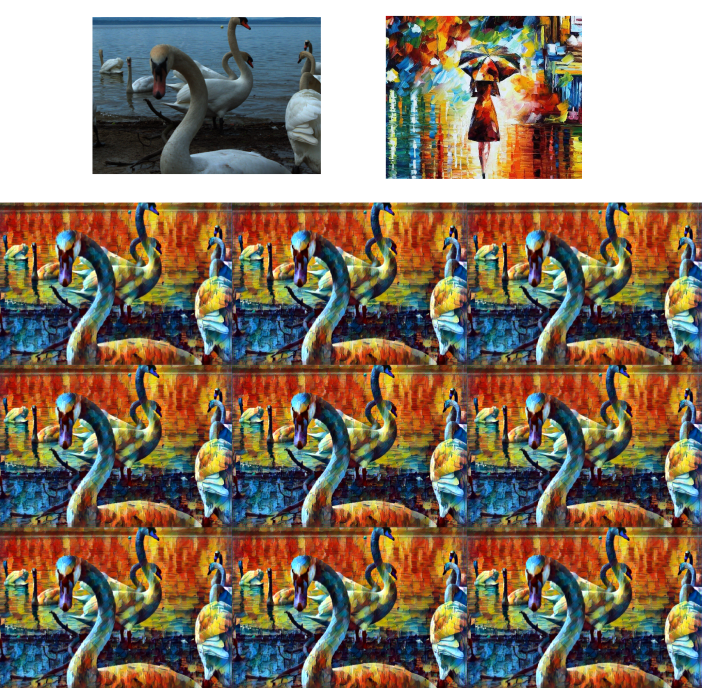}
\end{center}
   \caption{``Swans'' light field image stylized with our method. A subset of the full set of stylized views is shown. The views are selected with a stride of 3 to each side of the central view.
     \vspace{-0.10in}
}
\label{fig:SwansLF}
\end{figure}

\begin{figure}
\begin{center}
\begin{tabular}{cc}
\textbf{a)} &
\textbf{b)} \\
\includegraphics[width=.45\linewidth]{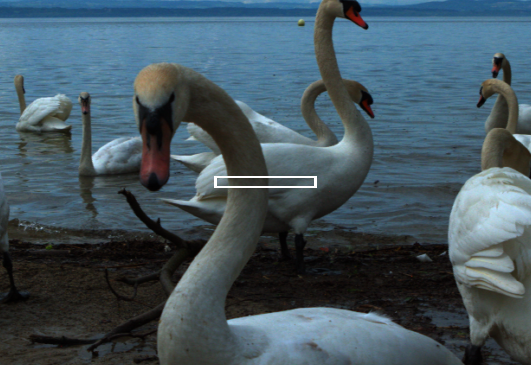}&
\includegraphics[width=.32\linewidth]{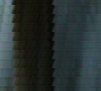}
\end{tabular}
\begin{tabular}{cc}
\textbf{c)} &
\textbf{d)} \\
\includegraphics[width=.45\linewidth]{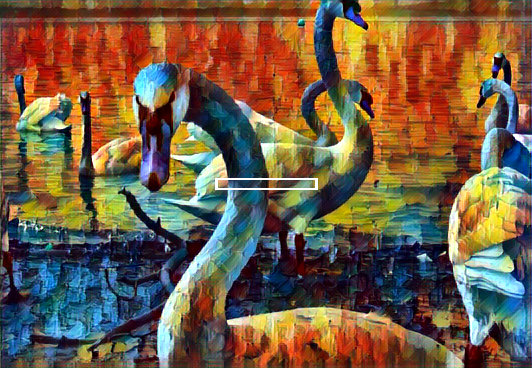}&
\includegraphics[width=.32\linewidth]{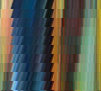}
\end{tabular}
\end{center}
   \caption{a) The central view of the ``Swans'' light field image. b) The epipolar image in the highlighted region part a. c) The central view of the stylized light field image. d) The epipolar image of the highlighted region part c.}
\label{fig:Epipolar}
\end{figure}

\subsection{Qualitative Evaluation}


Our method works for a variety of models and images. 
Fig.~\ref{fig:SwansLF} 
shows subsets of the views from light fields stylized with the proposed method. 
To more clearly see the consistency and shift between views, We also provide epipolar images of the stylized light fields in Fig.~\ref{fig:Epipolar}.
Additional results can be found in the supplemental materials accompanying this paper, which include a video that better shows the angular consistency between shifting subaperture views.

\begin{figure}[t]
\begin{center}
\footnotesize
\begin{tabular}{ccc}
    \includegraphics[width=0.95in]{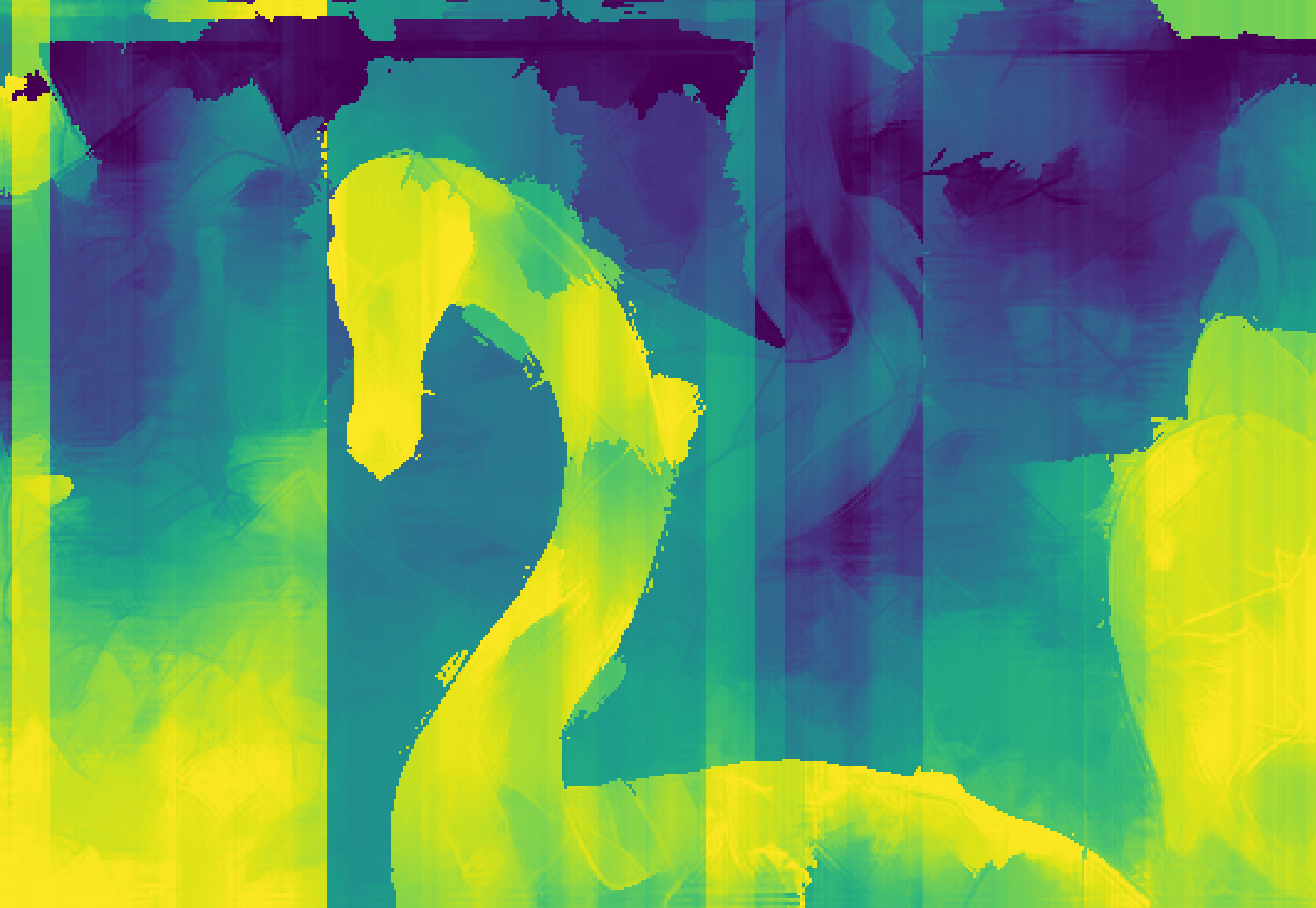} &
    \includegraphics[width=0.95in]{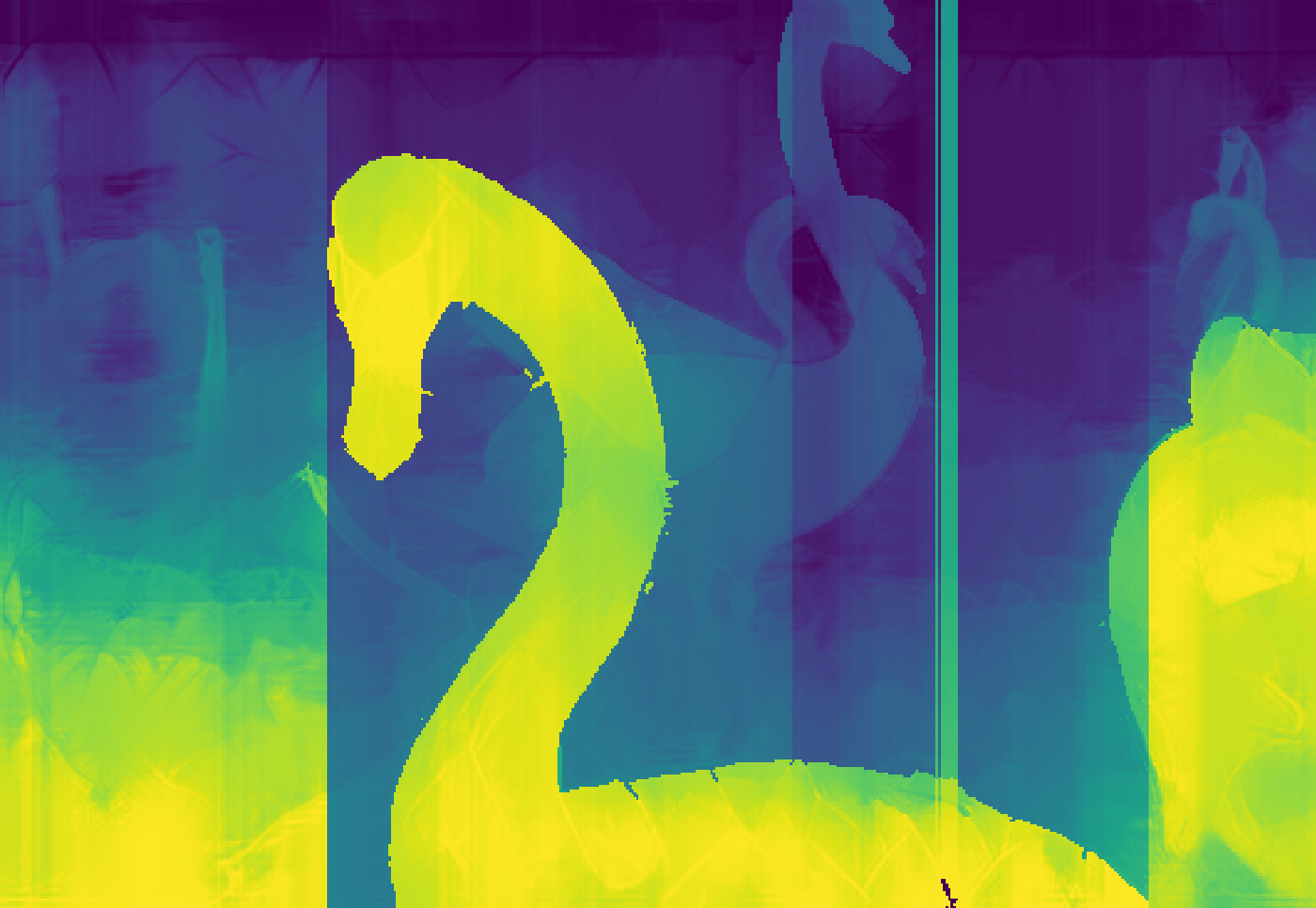} &
    \includegraphics[width=0.95in]{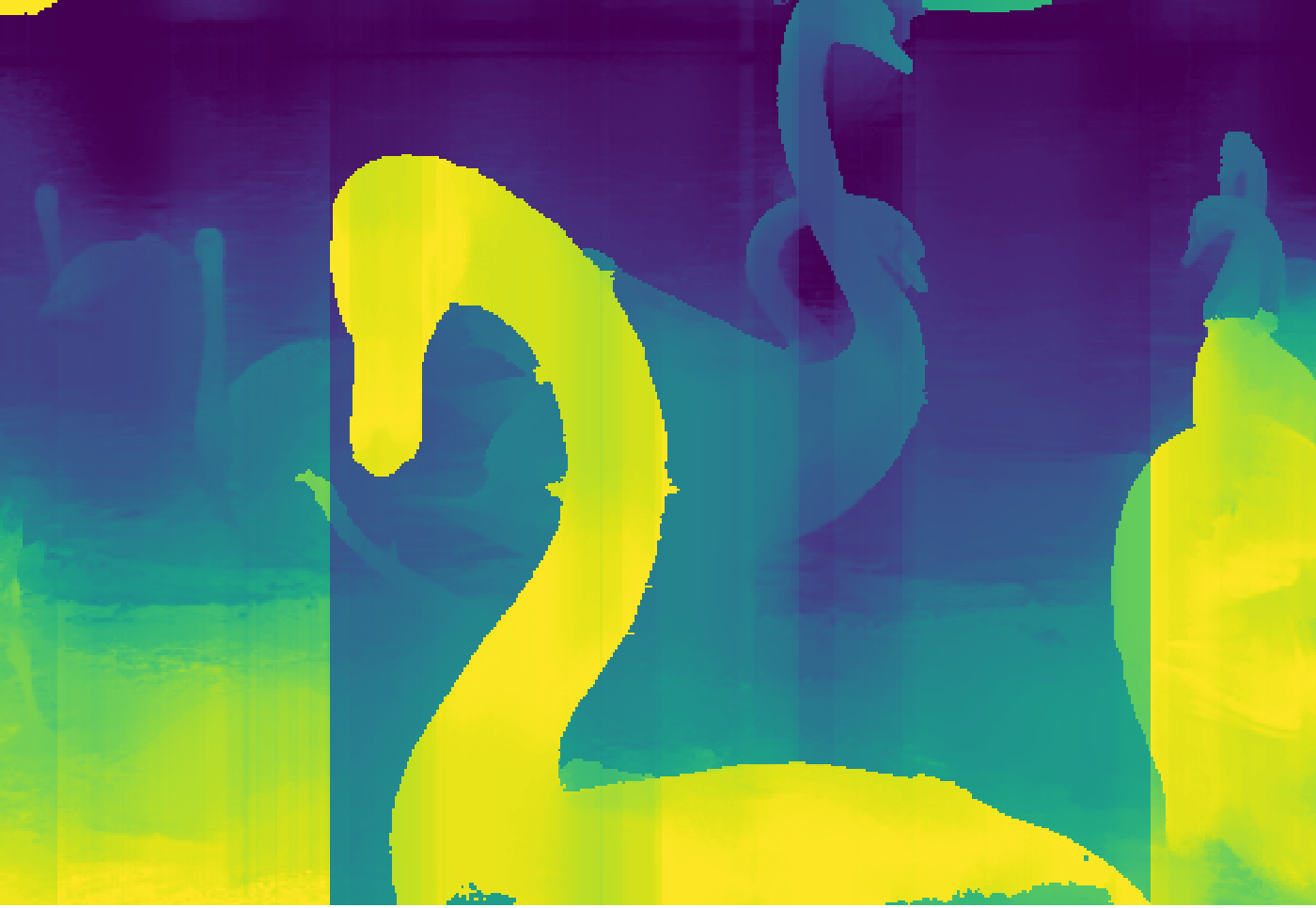} \\
    a) Naive & 
    b) Proposed & 
    c) Unstylized
\end{tabular}
\end{center}
   \caption{
   Depth maps computed~\cite{LightFieldDepth1} from stylized light fields.
   Depth maps computed from naively stylized light fields~(a) demonstrate errors due to the lack of angular consistency while those reconstructed from light fields stylized using our method~(b) are similar to those computed from original unstylized light fields~(c). 
   }
\label{fig:DepthRecomputation}
\end{figure}

In addition to visually inspecting the individual views for angular consistency, we can also determine how well the stylized light field maintains geometric properties of the original.
One way to verify this is to recompute depth maps from the stylized light fields and compare them to those of the unstylized light fields.
Fig.~\ref{fig:DepthRecomputation} shows an example of this using the ``Swans'' light field, the ``Mosaic'' style image, and the depth computation method from~\cite{LightFieldDepth1}.
As shown in Fig.~\ref{fig:DepthRecomputation}a, naively stylized light fields do a poor job retaining depth properties due to the lack of angular consistency.
Light fields stylized with our method do a better job preserving depth properties~(\ref{fig:DepthRecomputation}b) and are similar to those of the unstylized original~(\ref{fig:DepthRecomputation}c).

\begin{figure}
\begin{center}
\begin{tabular}{cc}
\textbf{a)} &
\textbf{b)} \\
\includegraphics[width=.45\linewidth]{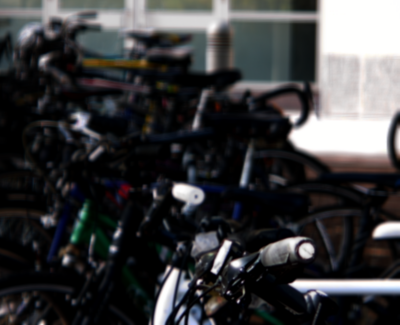}&
\includegraphics[width=.45\linewidth]{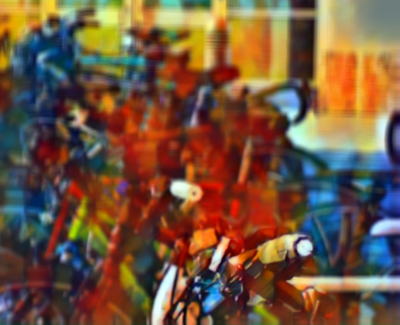}
\end{tabular}
\begin{tabular}{cc}
\textbf{c)} &
\textbf{d)} \\
\includegraphics[width=.45\linewidth]{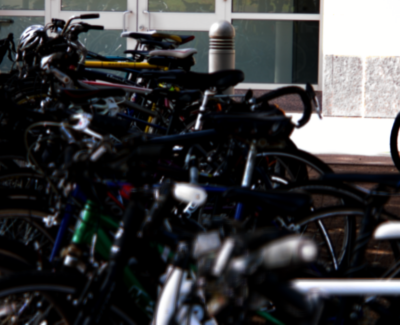}&
\includegraphics[width=.45\linewidth]{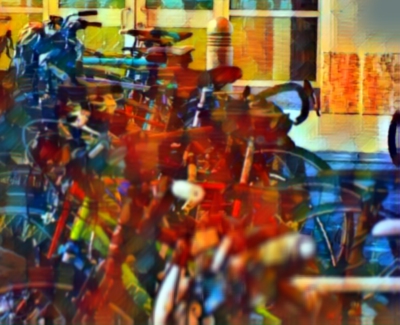}
\end{tabular}
\end{center}
   \caption{The focal stack of an example stylization. Near focus for the a)~original and b)~stylized light field. Far focus for the c)~original and d)~stylized light field.}
\label{fig:Refocus}
\end{figure}

Light field images are often used to render dynamically refocused images of the captured scene as first described in~\cite{NgDissertation}.
If the light field is angularly consistent, it should maintain the ability to refocus even in the stylized format.
This ability to refocus the stylized light field is demonstrated in Fig.~\ref{fig:Refocus}.

\subsection{Quantitative Evaluation}

The primary metric for neural style transfer is perceptual loss~\cite{FeedForwardStyleTransfer}, combining the ideas of content loss and style loss from~\cite{OriginalStyleTransfer}.
As in~\cite{StereoStyleTransfer2} and~\cite{StereoStyleTransfer1}, for multiview stylization we can combine this metric with disparity loss (Eq.~\ref{eq:disparity-loss}) to evaluate angular consistency.
An ideal light field stylization method should be able to minimize the disparity loss without increasing the perceptual loss.
In Table~\ref{table:NaiveVsOurs}, we compare our results to the baseline of naively stylizing each view independently.
This evaluation uses four different styles and presents the per-view average loss.
The proposed method causes only an extremely small increase in perceptual loss across all four styles, which is to be expected since this is balanced against disparity loss.

The most significant change is in the disparity (angular consistency) loss, which drops by an order of magnitude or more, quantitatively validating the visual consistency demonstrated in Figs.~\ref{fig:naive} and~\ref{fig:SwansLF}.

\begin{table}
\centering
\scriptsize 
\begin{tabular}{ |c|r|r|r|r| }
 \hline
    & \multicolumn{4}{|c|}{Perceptual Loss} \\
 \hline
  & \multicolumn{1}{|c|}{Candy} & \multicolumn{1}{|c|}{Mosaic} & \multicolumn{1}{|c|}{Rain Princess} & \multicolumn{1}{|c|}{Udnie} \\ 
  \hline
 Naive & 3954244 & 4520050 & 3628793 & 830148\\  
 Ours & 3954242 & 4520043 & 3628792 & 830148\\
 \hline
    & \multicolumn{4}{|c|}{Disparity Loss} \\
 \hline
 Naive & 6044 & 8807 & 3845 & 690\\  
 Ours  & 113 & 139 & 102 & 40\\ 
 \hline
\end{tabular}
\vspace{0.10in} 
 \caption{Evaluation of perceptual and disparity loss for multiple stylization models. Our method keeps similar perceptual loss to the naive method while greatly decreasing the disparity loss.
}
 \label{table:NaiveVsOurs}
\end{table}


\section{Variations and Experiments}
\label{sect:variations}


In addition to evaluating the proposed method, we also explore several variations and simplifications to evaluate the relative contributions of various elements of the approach.

As noted in Section~\ref{sect:full-method}, our approach begins with a pre-trained style transfer network and then iteratively optimizes the network variables to reduce the disparity loss for a single image rather than trying to train a single network to function in a purely feedforward way that generalizes to other images.
This raises the question of whether one could simply use a purely optimization-based approach such as a Gatys-like network that incorporates perceptual loss and the additional disparity loss term to encourage consistency.
We have explored that option and found that although it produces good results,
 the optimization does all of the work from scratch instead of being able to leverage a pre-trained stylization network and explicitly warped-and-fused feature maps.  
As such, it requires significantly more iterations and typically takes about twice as long to run as the proposed method.

We explore other variations of the full method proposed in Section~\ref{sect:full-method} and illustrated in Fig.~\ref{fig:BigPicture}.  
These variations are described in the following subsections and summarized in Table~\ref{table:variations}.
For comparison of the possible variations, we analyze the average per-view perceptual loss and disparity loss for a set of light fields, the results of which are given in Table~\ref{table:variations-eval}.

For the Naive method (independently stylized views), we again see that the disparity loss is high because no angular consistency was enforced. 
For a consistently stylized light field, we would expect the disparity loss to greatly decrease while the perceptual loss remains unchanged.
Table~\ref{table:variations-eval} shows that all of the variations explored here produce light fields with nearly identical perceptual loss. 
This is to be expected since any given view considered in isolation maintains the properties of the stylization, even if inconsistent with the other views. 
Thus, rather than focusing on perceptual loss, we use disparity loss as the main comparison metric when comparing and discussing the following variations.

\begin{table}[!b]
    \centering
    \scriptsize
    \begin{tabular}{|l|c|c|c|}
        \hline
        & Warp/Fuse Features & Warp/Fuse Images & No Fusion \\
        \hline
        Full BP & BPFuseFeatures$^1$ & BPFuseImg & BPNoFuse \\
        \hline
        Post-optimize & OptFuseFeatures & OptFuseImg & OptNoFuse \\
        \hline
        No iteration & NaiveFuse & WarpBlend & Naive$^2$ \\
        \hline
        \multicolumn{2}{l}{$^1$ The full method proposed in Section~\ref{sect:full-method}} \\
        \multicolumn{2}{l}{$^2$ What~\cite{StereoStyleTransfer2} refers to as ``baseline''}
    \end{tabular}
    \caption{Variations on the proposed method explored in Section~\ref{sect:variations}
}
    \label{table:variations}
\end{table}

\begin{table}
    \centering
    \scriptsize
    \begin{tabular}{|l|c|c|c|}
        \hline
        & \multicolumn{3}{|c|}{Perceptual Loss} \\
        \hline
        & Warp/Fuse Features & Warp/Fuse Images & No Fusion \\
        \hline
        Full BP & 3193389 & 3193389 & 3193388 \\
        \hline
        Post-optimize & 3193386 & 3193388 & 3193380 \\
        \hline
        No iteration & 3193392 & 3193389 & 3193391\\
        \hline
        & \multicolumn{3}{|c|}{Disparity Loss} \\
        \hline
         Full BP & 98 & 16 & 363 \\
        \hline
        Post-optimize & 15 & 14 & 14 \\
        \hline
        No iteration & 2089 & 25 & 4846 \\
        \hline
        & \multicolumn{3}{|c|}{Execution Time (seconds)} \\
        \hline
         Full BP & 308 & 304 & 304 \\
        \hline
        Post-optimize & 135 & 137 & 136 \\
        \hline
        No iteration & 43 & 45 & 42 \\
        \hline
   \end{tabular}
    \caption{Comparison of perceptual loss, disparity loss, and execution time for variations of the proposed method
    \vspace{-0.1in}
    }
    \label{table:variations-eval}
\end{table}

\subsection{Fusion Variations}
\label{sect:fusion-variations}

\begin{figure}
\begin{center}
\includegraphics[width=0.95\linewidth]{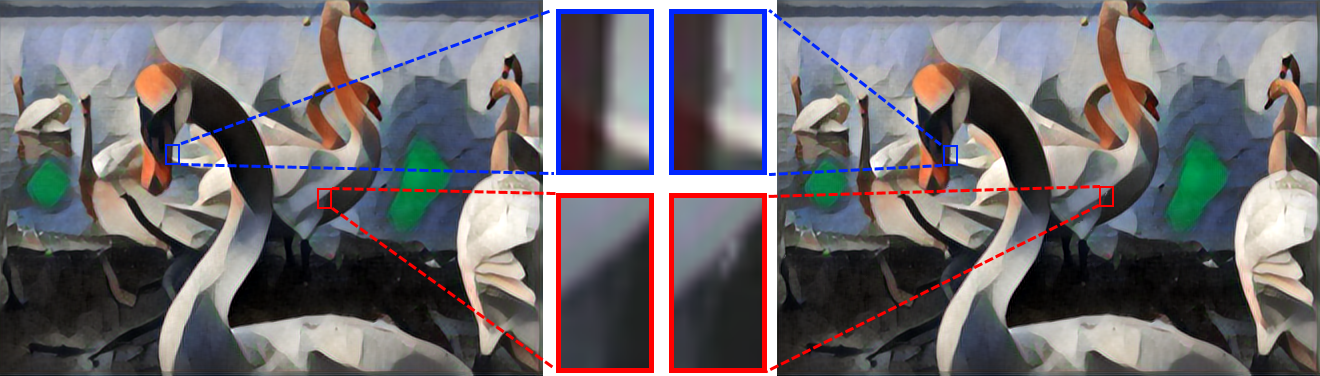}
\end{center}
    \caption{Two views of ``Swans'' stylized using the WarpBlend variant (image-space fusion without subsequent optimization). 
    Methods that fuse in the image domain are highly dependent on accurate disparity maps. 
    Errors in the disparity maps lead to visual artifacts in the final stylized light field, such as those shown in the red and blue callouts.
    }
    \label{fig:fuse-image-artifacts}
\end{figure}

Our method, like~\cite{StereoStyleTransfer2} and~\cite{StereoStyleTransfer1}, pairwise fuses elements of two images in the feature map domain.
This raises the question of whether such feature-map fusion is preferable to image-space fusion (i.e., fusing $I'_{0,0}$ and $I'_{s,t}$ instead of $F_{0,0}$ and $F_{s,t}$) or is even required at all.
When we analyze the methods that use fusion in the image domain (represented as the middle column in Table~\ref{table:variations-eval}), it is clear that these methods have the lowest masked disparity loss.
However, visual analysis of the stylized light fields produced with these methods shows that artifacts appear frequently in the unmasked regions, {\em which are not factored into the quantitative masked disparity loss}.
Warping and fusing in the image domain also relies heavily on the notion that this process is done with perfect disparity maps.
In reality, noise in the image data, featureless regions, ambiguous matches, discrete pixel sampling, and other factors cause imperfect depth or disparity estimates, all of which are well known issues with stereo, multi-view stereo, and light-field depth estimation.
This over-reliance on the accurate disparities can cause additional artifacts that are undesirable in the final images as shown in Fig.~\ref{fig:fuse-image-artifacts}.
We believe that fusing feature maps and then decoding them results in visually better (more artifact-free) stylizations than image-space fusion after decoding because the decoding of the feature maps mitigates such artifacts.

Since the fusion step takes time, we also consider whether fusion is even necessary for the optimization to converge and whether it could be discarded in order to save processing time.
While the disparity loss is comparable to that of other backpropagating methods, the lack of a fusion step causes the network to take longer to converge on each individual view, especially the earliest optimized views.
If this method is trained with the same learning rate, number of epochs, and optimization sequence as described in Section~\ref{sect:full-method}, it results in some views having ghosting artifacts, especially in areas with high frequency content.
Thus, the number of optimization epochs must be increased to produce results comparable to the proposed method, more than offsetting any potential time savings.

\subsection{Optimization Variations}

Our method performs optimization by backpropagating all the way back through the stylization encoder/decoder, essentially the same as extreme overfitting of the network to a single set of light-field views.  
Another option would be to use the feed-forward stylization network, including warping and fusion of the feature maps, to produce initial estimates of $I'_{s,t}$ and then post-optimize with the pixels of $I'_{s,t}$ as the only updated variables.
Such a method is essentially the same as generating the $I'_{s,t}$ images using warped and fused feature maps and then running it through a Gatys-like optimization including the additional disparity loss term.
It is also worth considering if a light field can be stylized without the need for an optimization at all, but relying solely on warping and fusing feature maps to create consistency.
Upon experimenting with these optimization variations, we find three key findings.

First, fusing features and then post-optimizing $I'_{s,t}$ to reduce disparity loss without backpropagating through the network (OptFuseFeatures) results in worse angular consistency than propagating the loss back through the stylization encoder/decoder (BPFuseFeatures, our primary method).
This can be attributed to the initial fusing of features $F_{0,0}$ with $F_{s,t}$, which essentially alpha-blends features maps from a network that has not trained on disparity loss.
Allowing the feed-forward stylization encoder/decoder to train on disparity loss allows these to learn to produce more consistent feature maps.

Second, the WarpBlend method, which uses image-space warping and blending with no subsequent optimization, gives reasonable results with a roughly 12x speedup compared to the  method proposed in this paper since it requires no iterative optimization.
This method essentially involves independently stylizing each view using the pre-trained stylization network and then employing a purely image-space approach to warp the stylized central view $I'_{0,0}$ to each of the other views and blending it with the initial stylization $I'_{s,t}$ for those views using the correspondence confidence map $M_{s,t}$.
This means it could serve as an alternative to the proposed method if greater speed is desired.
However, we reiterate that because it relies {\em entirely} on accurate disparity-based warping in image space, it is susceptible to qualitative visual artifacts from disparity errors as discussed earlier in Section~\ref{sect:fusion-variations} and shown in Fig.~\ref{fig:fuse-image-artifacts}.
These are not factored into the masked disparity loss.

Third, we find all methods of post-optimization on the output image to be undesirable. 
Since it can only optimize on the disparity loss, it eventually converges to give essentially the same results as the WarpBlend method (which can be thought of as the minimization of the disparity loss), and thus retains all the same visual artifacts. 



\subsection{Loss Function}

\begin{table}[b]
    \centering
    \scriptsize
    \begin{tabular}{|l|c|c|c|}
        \hline
        & \multicolumn{3}{|c|}{Perceptual Loss} \\
        \hline
        & BPFuseFeatures & BPFuseImg & BPNoFuse \\
        \hline
        Both Loss Terms &  3193389 & 3193389 & 3193388 \\
        Disp. Loss Only & 3193389 & 3193389 & 3193388 \\
        \hline
        & \multicolumn{3}{|c|}{Disparity Loss} \\
        \hline
        & BPFuseFeatures & BPFuseImg & BPNoFuse \\
        \hline
        Both Loss Terms & 98 & 16 & 363 \\
        Disp. Loss Only & 99 & 16 & 366 \\
        \hline
        & \multicolumn{3}{|c|}{Execution Time (seconds)} \\
        \hline
        & BPFuseFeatures & BPFuseImg & BPNoFuse \\
        \hline
        Both Loss Terms & 566 & 565 & 565 \\
        Disp. Loss Only & 308 & 304 & 304 \\
        \hline
   \end{tabular}
    \caption{Comparison of backpropagating / optimizing using combined perceptual and disparity loss to using disparity loss alone}
    \label{table:full-vs-disp-loss}
\end{table}

\begin{figure}
\begin{center}
\includegraphics[width=0.75\linewidth]{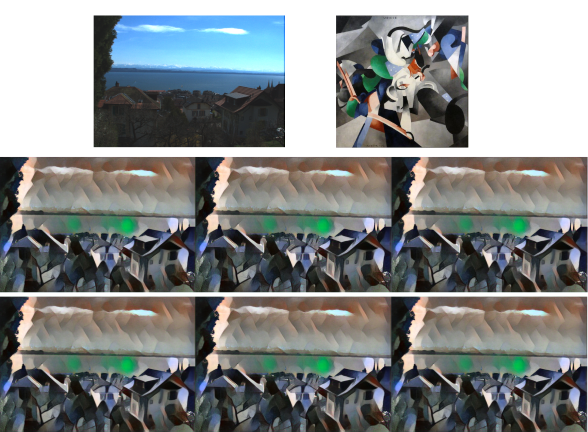}
\vspace{-0.1in}
\end{center}
    \caption{``Lake'' light field stylized using BPFuseFeatures optimized with the disparity loss and perceptual loss (top) and the disparity loss only (bottom). Results are visually indistinguishable. }
    \vspace{-0.05in}
    \label{fig:disp-vs-full}
\end{figure}

\begin{figure}
\begin{center}
\includegraphics[width=0.68\linewidth]{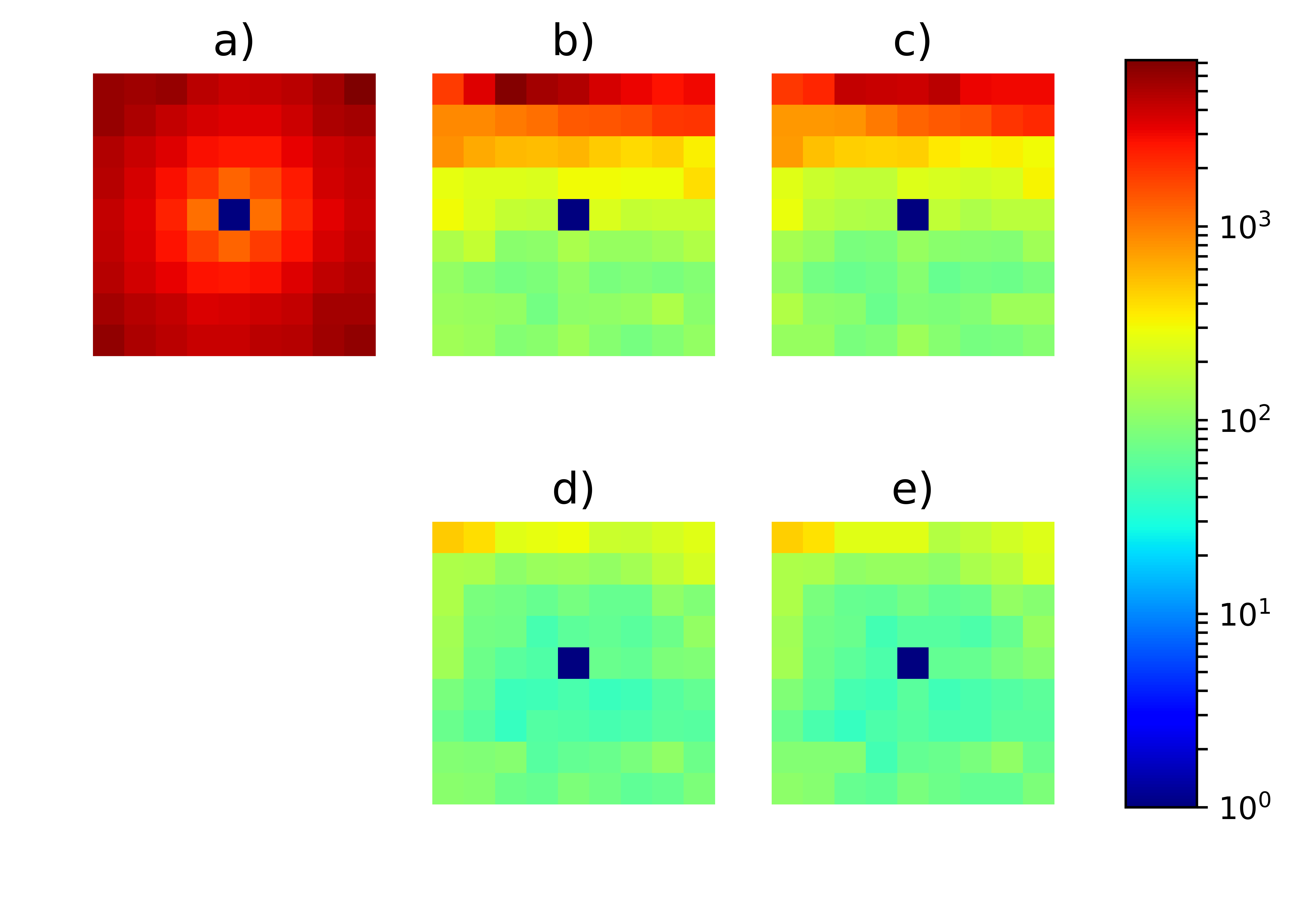}
\vspace{-0.25in}
\end{center}
    \caption{Visualization of disparity loss by view for 
    a)~Naive, 
    b)~BPNoFuse optimized on disparity loss only, 
    c)~BPNoFuse optimized on disparity and perceptual loss, 
    d)~BPFuseFeatures optimized on disparity loss only, 
    and 
    e)~BPFuseFeatures optimized on disparity and perceptual loss. 
    \vspace{-0.05in}
}
    \label{fig:loss-graphs}
\end{figure}

For the results described so far in this section, we either backpropagate through the stylization network or post-optimize the output image using only disparity loss.
Given that the stylization network has been pre-trained to minimize perceptual loss~\cite{FeedForwardStyleTransfer}, we consider the question of whether it is effective to include perceptual loss to minimize visual artifacts along occlusion boundaries.
However, we have found that backpropagating perceptual loss and disparity loss produces results that are nearly indistinguishable visually from those created using disparity loss alone, as shown in Fig.~\ref{fig:disp-vs-full} and in Table~\ref{table:full-vs-disp-loss}.
This is also evident when analyzing the disparity loss for each view of the light field as shown in Fig.~\ref{fig:loss-graphs}.
Excluding perceptual loss from the optimization also avoids having to backpropagate through the VGG-16 network at the end of the overall network, reducing the computation required.
We have found that this reduces the execution time by approximately 35\% with comparable results.

\section{Conclusion}

This paper presents the first neural style-transfer method for light fields that achieves high-quality visual results while maintaining angular consistency. 
The method uses a given central-view depth map to create masked and confidence-weighted disparity maps for each other view, allowing backward warping from the central view to all other views. 
This warping and masking is vital to the optimization process and consistently stylized results. 
As with recent methods for stereoscopic neural stylization, we fuse warped feature maps using a confidence-weighted mask.  
Unlike these methods, 
we do not try to train a single network to stylize different light fields in a purely feed-forward fashion.
Instead, we incorporate pre-trained monocular style-transfer networks and iteratively optimize them for each view.

These results are validated both qualitatively (visually) and quantitatively.
We also present variants of this method that allow for trade-offs between angular consistency, sensitivity to errors in the original depth map, and execution time.

{\small
\bibliographystyle{ieee}
\bibliography{egpaper_final}
}

\end{document}